\pdfoutput=1

\documentclass[11pt]{article}

\usepackage[final]{acl}
\usepackage{times}
\usepackage{latexsym}
\usepackage{booktabs, makecell, longtable}
\usepackage{hyperref}
\usepackage{amsmath}
\usepackage{graphicx}
\usepackage{paralist}
\usepackage{amssymb}
\usepackage{lipsum}
\usepackage[width=.90\textwidth]{caption}
\usepackage[T1]{fontenc}

\usepackage[utf8]{inputenc}

\usepackage{microtype}

\title{Improving the Generalizability of Collaborative Dialogue\\ Analysis With Multi-Feature Embeddings}

\author{Ayesha Enayet \\
University of Central Florida \\
Orlando, FL USA\\
\texttt{ayeshaenayet@knights.ucf.edu} \\\And
  Gita Sukthankar \\
 University of Central Florida\\
Orlando, FL USA \\
  \texttt{gitars@eecs.ucf.edu} \\}

\begin{document}
\maketitle
\begin{abstract}
Conflict prediction in communication is integral to the design of virtual agents that support successful teamwork by providing timely assistance. The aim of our research is to analyze discourse to predict collaboration success. Unfortunately, resource scarcity is a problem that teamwork researchers commonly face since it is hard to gather a large number of training examples. To alleviate this problem, this paper introduces a multi-feature embedding (MFeEmb) that improves the generalizability of conflict prediction models trained on dialogue sequences. MFeEmb leverages textual, structural, and semantic information from the dialogues by incorporating lexical, dialogue acts, and sentiment features. The use of dialogue acts and sentiment features reduces performance loss from natural distribution shifts caused mainly by changes in vocabulary. 
 
This paper demonstrates the performance of MFeEmb on domain adaptation problems in which the model is trained on discourse from one task domain and applied to predict team performance in a different domain.   The generalizability of MFeEmb is quantified using the similarity measure proposed by \citet{bontonou2021predicting}.  Our results show that MFeEmb serves as an excellent domain-agnostic representation for meta-pretraining a few-shot model on collaborative multiparty dialogues.
\end{abstract}

\section{Introduction}
For many natural language processing applications, the ability to learn features that generalize well across multiple datasets is a key desideratum~\cite{saikia2020optimized}. 
This paper introduces a new multi-feature embedding, MFeEmb, that increases the generalizability of models learned from collaborative multiparty dialogues. Dialogues are different from single-author documents in that, along with textual information, they contain communication patterns that may serve as indicators of social dynamics.  Treating a dialogue as a mere text collection ignores valuable information. We advocate exploiting implicit features present in multiparty dialogues that are less vulnerable to distribution shifts resulting from task domain changes.

This paper demonstrates the usage of MFeEmb on a communication analysis task: conflict prediction.  Teamwork research faces a challenge of resource scarcity since the human subjects datasets are quite small (less than 100 samples), due to the difficulty of recruiting teams and the time consuming nature of many group tasks.  A variety of social phenomena have been investigated within team communication research including entrainment~\cite{rahimi2020entrainment2vec} and emergent leadership~\cite{emergentleadership}. Frequency of communication is not in itself a good predictor of team performance, but a meta-analysis conducted by \citet{marlow2018} that drew upon data from 150 studies conducted on 9702 teams concluded that high quality communication is positively related to team performance in many task domains.  Conversely, process conflict, “disagreement among group members about the content of the tasks being performed, including differences in viewpoints, ideas, and opinions”~\cite{jehn1995}, is usually negatively correlated with taskwork success. 

Our aim is to be able to learn a model to classify process conflict from multiparty dialogues that generalizes well across multiple tasks. We treat the task of conflict prediction as a binary classification task with high conflict and low conflict being the two classes; the ground truth used by the conflict prediction model is measured using a post-task team process conflict survey.  This paper focuses on three collaborative problem-solving tasks: software engineering, search and rescue, and cooperative gameplay.

Our proposed embedding, MFeEmb, leverages textual, structural, and semantic information from the dialogues by incorporating vocabulary, dialogue acts, and sentiment features. Lexical embeddings such as word2Vec and BERT~\cite{devlin2018bert} show good performance across multiple NLP tasks on in-domain test sets but are less robust to domain shift. 
Previous work identified that dialogue acts and sentiment sequences are informative features that predict conflict reliably even at the earliest stage of team problem-solving ~\cite{enayet2021analyzing}; however, classifiers constructed using these features still experience lackluster transfer performance when applied to new datasets, particularly when detecting high conflict examples~\cite{enayet2021learning}.

To address this transfer problem, we propose the usage of MFeEmb, specifically as a meta-pretraining representation to be used within a few-shot model.  MFeEmb combines the strengths of both domain-invariant and domain-specific features.  This paper compares the generalizability potential of the MFeEmb embedding vs.\ standard word embeddings using inter-class and intra-class based similarity measures, proposed by \citet{bontonou2021predicting}. Then we evaluate the performance of MFeEmb in a domain adaptation scenario in which the model is trained on discourse from one task domain and used to predict conflict in a different domain. Our results show that:
\begin{compactenum}
\item MFeEmb demonstrates superior generalizability over other embeddings for collaborative multiparty dialogues. 
\item MFeEmb is an excellent representation choice for the meta-training stage of few-shot learning.
\item The domain adaptation performance of MFeEmb can be easily enhanced by task specific synonym replacement.
\end{compactenum}
\section{Related Work}
Previous studies on group interaction tasks such as conflict prediction~ \cite{rahimi2020entrainment2vec}, disruptive talk detection \cite{park2022disruptive}, group satisfaction~\cite{lai2018predicting}, and task performance prediction~\cite{kubasova2019analyzing,murray2018predicting} have focused on simply improving performance on in-domain datasets. Very little attention has been paid to the problem of creating generalizable models for multiparty dialogue that can be used when training data is scarce. The intelligent tutoring system community has empirically assessed the generalizability of common natural language representations, such as BERT and Linguistic Inquiry Word Count (LIWC), across collaborative problem solving tasks~\cite{pugh2022speech}, but without investigating methods to improve generalizability.


In domain adaptation, the goal is to train a model on data from a source domain that performs well on a test dataset drawn from a different target distribution.  Common NLP tasks (e.g., part-of-speech (POS) tagging and named entity recognition (NER)) have been tackled using techniques including instance weighting~\cite{jiang-zhai-2007-instance} or explicitly identifying feature correspondences between the domains ~\cite{blitzer2006}. An alternate approach is to learn a single representation that generalizes well across multiple domains.  This can be done using few-shot learning~\cite{fewshot},  one of the most widely used approaches to dealing with resource scarcity. The traditional framework comprises meta-training and meta-testing phases, where the aim of meta-training is to learn universal representations from multiple domains. 

\citet{triantafillou2021learning} introduced a method that improves few-shot generalizability by making use of multiple datasets in order to learn a universal template.  \citet{dvornik2020selecting} proposed Selecting from Universal Representations (SUR), which involves learning a multi-domain representation by training multiple feature extractors. A multi-domain feature bank is used to select the most relevant feature during the learning phase. Rather than seeking to learn the new representation entirely from data, our research exploits similarities in dialogue act sequences and sentiment patterns commonly observed during successful collaborative problem-solving.   



 Representation choice has been shown to  place an upper bound on target domain performance~\cite{generalizability}. 
 Few-shot frameworks such as Meta-pretraining then Meta Learning (MTM)~\cite{mtm} have assumed that word embeddings like BERT that are trained on large datasets are the best choice for task agnostic pre-training.  \citet{bontonou2021predicting} introduced a method to quantify the generalizability of a few-shot classifier under supervised, unsupervised and semi-supervised settings.  This paper uses their inter-class and intra-class based generalizability measure to evaluate MFeEmb vs.\ simple word-based embeddings under supervised classification scenarios. Our research demonstrates that MFeEmb is superior to word embeddings as a meta- pretraining representation.

\section{Methodology} This section describes our approach to learning a generalizable embedding from multi-party dialogues for conflict prediction.  We discuss our datasets, introduce our embedding, and show how our technique can be used in combination with data augmentation and few-shot learning.
\subsection{Datasets} Datasets collected from different collaborative problem-solving task domains were used in our study of generalizability:
\begin{compactenum}
\item \textbf{Teams corpus} \cite{litman2016teams}: this dataset consists of dialogue from 62 teams playing a cooperative board game in groups of three or four.  Each team plays the game twice together. The Teams corpus was originally created to study entrainment, a linguistic phenomena in which teammates adopt similar speech patterns~\cite{rahimi2020entrainment2vec}. The Game1 dataset of Teams corpus contains 62 dialogues, 32 low conflicts, and 30 high conflict dialogues. The Game2 dataset of Teams corpus contains 62 dialogues, 33 low conflicts, and 29 high conflict dialogues.   
\item \textbf{ASIST dataset} \cite{ASU/BZUZDE_2022}: This dataset consists of 67 teams of three people participating in a simulated search and rescue task within the Minecraft game environment.  Participants completed two different missions that involved searching a map and triaging victims.  The dataset was collected by the ASIST project to stimulate the development of proactive assistant agents for helping human teams. The dataset contains 113 dialogues, 58 low conflicts, and 55 high conflict dialogues. 
\item \textbf{GitHub social coding dataset}~\cite{github}: This dataset was mined directly from the GitHub social coding platform. It consists of data from issue comments of teams developing open source software over a period of months. Teams vary in size, and comments were harvested for 50 reported issues. The dataset contains 50 dialogues, 29 low conflicts, and 21 high conflict dialogues.
\end{compactenum}
Both the Teams and ASIST datasets contain post-task process conflict survey data for all teams, which we divide into high and low conflict groups using their z-scores.
 For GitHub, process conflict was scored according to an issue resolution rubric  (described in Appendix \ref{sec:github_scoring}). 
\subsection{Multi-Feature Embedding (MFeEmb)} \label{multFe} This paper introduces the MFeEmb embedding which is designed to capture the dialogues' structural, semantic, and textual information for collaborative task success prediction. To represent the structural information, we incorporate information from dialogue acts (DAs) of the utterances. For semantics, the sentiment polarities of the utterances are used, although DAs capture both semantic and structural information. Textual information is extracted from the vocabulary of the dialogues. 

 For the word embedding, we use both the Distributed Bag of Words and Dynamic Memory models of Doc2Vec~\cite{docvec} to learn embeddings (see Appendix \ref{App_doc2vec}). 
Although there is only 28\% vocabulary overlap between the ASIST and Teams datasets and 35\% overlap between the GitHub and Teams datasets, word embeddings can help preserve high performance on the training dataset, while including structural and semantic features makes the embedding more robust to domain shifts. 

For the dialogue act (DA) embedding, we first map the sequence of utterances to a sequence of DAs using USE-DAC (Universal Sentence Encoder Dialogue Act Classifier, described in Appendix \ref{App_DAC}).  The SwDA-DAMSL tagset was used to categorize dialogue acts. The TextBlob python module was used to assign sentiment polarities ranging from -1 to 1 to each of the utterances.

To generate the embeddings, we use the Dynamic Memory model of Doc2Vec due to the small vocabulary size of the sequences, which is limited by the number of DA tags and sentiment gradations. The Dynamic Memory model leverages context when generating embeddings, thus preserving information contained in these communication patterns. In contrast, the Distributed Bag of Words model does not consider context when generating embeddings. For the few-shot results, we also report results with pre-trained Word2Vec embeddings. First, we separately learn three embeddings from the sequence of DAs, sentiments, and utterances (text); the final MFeEmb embedding is created either by concatenating the three embeddings or by using LSTMs to learn a concatenation ensemble model. We have made our code available at \url{ https://github.com/ayeshaEnayet/MFeEmb.git}.
 \subsection{Corpus-Based Feature Analysis}  
\begin{figure*}
    \centering
    \captionsetup{width=.9\textwidth}
    \includegraphics[width=0.9\textwidth,height=4cm] {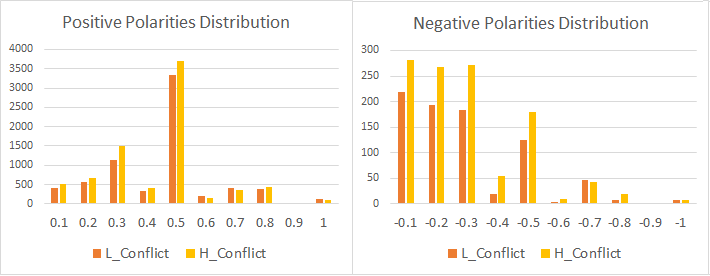}
    \caption{Sentiment polarity distribution of the high conflict vs. low conflict classes in the Teams dataset}
    \label{fig:Senti_Distribution}
\end{figure*}

 To understand the ramifications of our feature selections, we performed frequency distribution analyses across the high conflict and low conflict classes of the Teams Dataset. This analysis shows that the high conflict class has a high frequency of negative sentiment polarities compared to the low conflict class and a comparable frequency of positive sentiment polarities compared to the low conflict class (Figure~\ref{fig:Senti_Distribution}).
 
 In the dialogue act distribution, Statement-non-opinion (sd) is the most frequent tag in both classes. The low conflict class has a high frequency of positive communication indicators like Appreciation (ba), Conventional-closing (fc), and Thanking (ft) compared to the high conflict class.  The high conflict class contains a high frequency of bad communication indicators like Uninterpretable (\%), Hedge (h), Signal-non-understanding (br), and Apology (fa). Interestingly, high conflict classes have a high frequency of all categories of questions compared to low conflict classes (see dialogue act distributions and n-grams in Appendix~\ref{sec:freq_distrib}).

 Looking at the vocabulary distribution, the high conflict class contains more profanity words than the low conflict class, and there is no overlap between the profanity word lists of both classes.  Our analysis reveals that there is value in all three types of features (dialogue acts, sentiment polarity, and vocabulary) but that conflict prediction remains a challenging classification problem.
\subsection{Synthetic Datasets}
\label{sec:syn}
To further improve generalization, we augment our training data with synthetic datasets generated using synonym replacement, as proposed by \citet{wei-zou-2019-eda}.
Our data augmentation strategies are described below: 
\begin{compactenum}
    \item \textbf{SynReplace}: We augment Teams Game1 and Game2 by replacing the words with synonyms drawn from WordNet.
    \item \textbf{ASISTReplace}: We augment Teams Game1 and Game2 by replacing the words with only the synonyms present in the ASIST dataset. First, we extract the vocabulary of the ASIST dataset. During the replacement operation, we search for synonyms in WordNet and only replace them with the synonyms present in the ASIST dataset's vocabulary.
    \item \textbf{GitReplace}: Similar to ASISTReplace, we generate our third dataset by replacing the words with only the synonyms present in the GitHub dataset.
\end{compactenum}
 Four synthetic dialogues are generated for each dialogue of the Teams dataset after applying random replacement on 10\% of the words.  Our intuition is that collaborative problem-solving domains such as software engineering may contain a lot of task specific jargon, and even simple synonym replacement techniques greatly facilitates generalization.

\begin{figure}
    \centering
    \captionsetup{width=.9\linewidth}
    \includegraphics[width=0.4\textwidth]{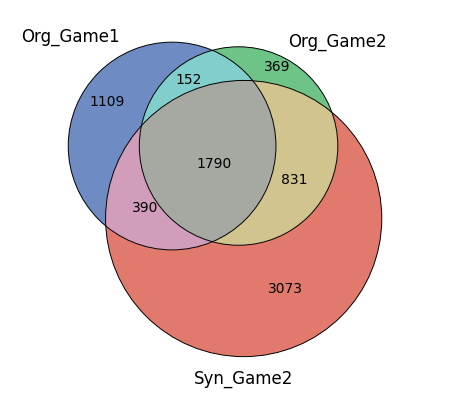}
    \caption{Vocabulary overlap between original Game1, original Game2, and the Game2 adversarially generated dataset}
    \label{fig:ven_diagram}
\end{figure}

In our experiments, the basic synonym replacement did not significantly change the intent and sentiment of the utterances.  To show the robustness of dialogue acts and sentiment sequences towards data augmentation, we utilize TextAttack~\cite{morris2020textattack}, a python package for adversarial attack and data augmentation, to generate a Teams Game2 synthetic dataset.  Word Swap by BERT-Masked LM transformation was employed to generate synthetic examples from the Teams Game2 dataset. One synthetic example is generated per dialogue of the Game2 dataset. The synthetic dataset contains $\approx 50\%$  more unique words than the original Game2 dataset (Figure \ref{fig:ven_diagram}).
Hamming distance was used to calculate the difference between the sequences of the Game2 original and Game2 synthetic datasets.   On average, the adversarial synthetic dataset only resulted in a 11\% change in DA sequences and 14\% change in sentiment sequences (Appendix~\ref{app:adversary}). 
\section{Experimental Setup}
 The Teams corpus contains 124 team dialogues from 62 different teams, playing two different collaborative
board games. We use the Teams Game1 dataset with 62 total samples, divided into 32 low and 30 high conflict samples, as our training dataset.  The small training dataset ensures that the experiments reflect the generalization performance under the resource scarcity scenario.   Our test datasets for evaluating domain adaptation are Teams Game2, GitHub, and ASIST. Obviously the domain shift is the smallest between the Teams Game 1 and 2 datasets.
 We use the GitHub and ASIST datasets to check the transferability of MFeEmb under domain shift. The model was not fine-tuned before evaluating the performance on GitHub and ASIST.

We evaluate our proposed MFeEmb under the following three experimental setups:
\begin{compactenum}
    \item SVM and logistic regression classifiers to distinguish high conflict and low conflict classes.
    \item LSTM concatenation ensemble. 
    \item Few-shot learning approach.  
\end{compactenum}
We benchmark MFeEmb against prior work on conflict prediction, other embedding choices, and FsText, a few-shot model proposed by \citet{bailey2018few}. Experiments were performed using a 300-dimensional version of MFeEmb where the length of all the three embeddings is the same, i.e., 100. We report the mean and standard devication of F1-Scores after 15 runs. For the SVM and logistic regression classification experiments, to improve the readability, we only report the best-performing classifier results measured by mean F1 score in Section \ref{results_sec}; for the results of both classifiers, see Appendix \ref{sec:svmLogresults}.  `*' denotes that logistic regression was the top performer, and `+' denotes cases where the SVM was the best.
\subsection{SVM and Logistic Regression} \label{sec:SVMLOG}
After Doc2Vec is used to generate the three embeddings for each sample, the embeddings are concatenated to create MFeEmb. We use both SVM and logistic regression to classify the instances and report the results of both the classifiers. For DAs and sentiment sequences, we always use the Dynamic Memory model (DM) of Doc2Vec. 
\subsection{Few-Shot Learning (FsText)} 
For few-shot learning, we use the method proposed by \citet{bailey2018few} and available in the FsText Python module. 
The training document for the meta-training stage of few-shot learning is represented using a pre-trained word embedding (Word2Vec).  In the case of more than one training sample per class, the proposed method works by averaging each class's vectors to calculate the most effective class representative. Cosine similarity is used to measure the distance between the test sample and each class representative, and the test sample is assigned the label of the class with the highest similarity.  We compare the generalizability of FsText (Original) with MFeEmb-based FsText, by replacing Word2Vec embedding with MFeEmb during the meta-training stage.
\subsection{Concatenation Ensemble} \label{sec:LSTM} Due to the small size of the training set, we apply the synonym replacement technique proposed by \citet{wei-zou-2019-eda} to augment the training data as described in Section~\ref{sec:syn}.  One hot encoding is used to encode DA, sentiment polarities, and vocabulary to train the model. We train three different Bidirectional LSTM models, one on each of DAs, sentiments, and word-based documents, and merge them to create our MFeEmb based ensemble.  
Our Bidirectional LSTM models for each feature have an embedding layer, an LSTM layer, one dropout layer, and one deep layer. 
\subsection{Baseline Models} We compare our proposed MFeEmb's results with several baseline models that use the same binary classification setup for conflict prediction.  
First, we show that MFeEmb performs competitively against prior work on conflict prediction~\cite{enayet2021analyzing} using the proposed dialogue act only  and sentiment only embeddings.  Note that our results are not directly comparable to what was reported in the previous work because we use a reduced training set; thus, we reimplemented the embeddings. We also compare MFeEmb to the commonly used BERT based embedding (Appendix~\ref{sec:bert}).

These independent baselines are compared against three implementation options for MFeEmb: 1) MFeEmb with simple binary classifier (SVM or logistic regression), 2) MFeEmb concatenation ensemble learned with LSTMs (Sec.~\ref{sec:LSTM}) trained on the synonym replaced augmented dataset, 3) a variation of few-shot learning method (FsText)~\cite{bailey2018few} in which the Word2Vec embedding is replaced with MFeEmb during the meta-training stage.  For training and testing, we concatenate all the utterances of the dialogue into one single document and assign it to one of the classes depending on the conflict score of the team. 
\section{Results} \label{results_sec}
\begin{figure*}[!h]
    \centering
    \includegraphics[width=0.95\textwidth,height=3.5cm]{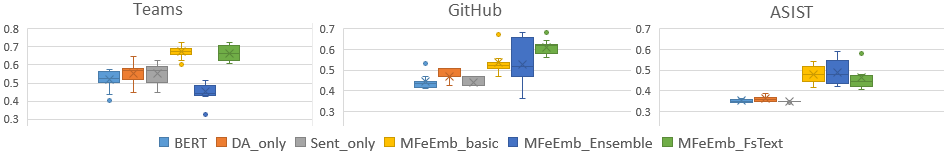}
    \caption{Performance of MFeEmb vs. other embedding choices from prior work.} 
    \label{fig:TableForDRGita_2}
\end{figure*}
\begin{figure*}[!h]
    \centering
    \includegraphics[width=0.95\textwidth,height=3.5cm]{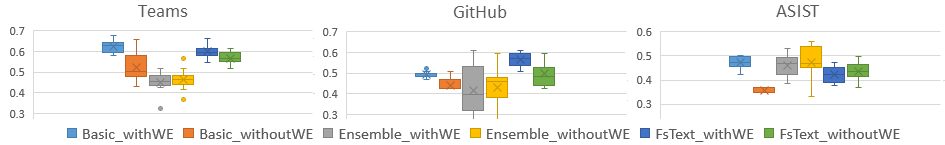}
    \caption{Performance of MFeEmb with and without word embedding (WE).}
    \label{fig:ImportanceOfWordEmb}
\end{figure*}
This section presents results on the generalizability of MFeEmb under different experimental setups.
\subsection{Similarity Based Evaluation} 
First we quantify the potential generalization of the representation using the similarity measure proposed by \citet{bontonou2021predicting}.  The similarity measure is given by: 
\begin{equation}
    intra(c)=\frac{1}{k(k-1)}\sum_{\substack{i \\ y_i=c}}\sum_{\substack{j\neq i \\ y_j=c}}\cos{(f_i,f_j)}
\end{equation}
\begin{equation}
    inter(c,\tilde{c})=\frac{1}{k^2}\sum_{\substack{i \\ y_i=c}}\sum_{\substack{j\neq i \\ y_j=\tilde{c}}}\cos{(f_i,f_j)}
\end{equation}
\begin{equation}
    similarity=\frac{1}{N}\sum_{c=1}^{N} (intra(c)- \max\limits_{c\neq \tilde{c}}( inter(c,\tilde{c})))
\end{equation}
where $c$ is class, $N$ is the number of classes, $k$ is number of examples, $f$ is the embedding, $intra(c)$ is cosine similarity within a class, and $inter(c,\tilde{c})$ is cosine similarity through classes $c$ and $\tilde{c}$. The final similarity score reflects the comparison  of the $intra(c)$ and $inter(c,\tilde{c})$. 
Intuitively it can be seen that the score measures how the representation affects the data clustering within and between classes. 

We compare our proposed MFeEmb vs.\ a standard word embedding learned using the bag of word model of Doc2Vec.  Table \ref{similarity} gives the result of the similarity-based analysis, juxtaposed with the classification results. MFeEmb has a better similarity score and high classification performance, compared to word-based embeddings indicating the high generalizability potential of MFeEmb.  
\begin{table}[!h]
\centering
\captionsetup{width=\linewidth}
\begin{tabular}{|p{1.5cm}|p{1.5cm}|p{1.5cm}|p{1.5cm}|}
\hline
\multicolumn{2}{|c|}{\textbf{\small Word\_{Emb}}}&\multicolumn{2}{|c|}{\textbf{\small MFeEmb}}\\
\hline
\multicolumn{4}{|c|}{\textbf{\small Teams Game2}}\\
\hline
\small similarity&\small F1\_score&\small similarity&\small F1\_score\\
\hline
\small -0.067&\small 0.470*&\small -0.016&\textbf{\small 0.628+}\\
\hline
\multicolumn{4}{|c|}{\textbf{\small GitHub}}\\
\hline
\small similarity&\small F1\_score&\small similarity&\small F1\_score\\
\hline
\small -0.067&\small 0.463*&\small -0.017&\textbf{\small 0.501+}\\
\hline
\multicolumn{4}{|c|}{\textbf{\small ASIST}}\\
\hline
\small similarity&\small F1\_score&\small similarity&\small F1\_score\\
\hline
\small -0.067&\small 0.446+&\small -0.016&\textbf{\small 0.458+}\\
\hline
\end{tabular}
\caption{Similarity-based generalizability analysis.  \textbf{Word\_Emb}: Distributed Bag of Words  document embedding. \textbf{MFeEmb}: Multi-Feature Embedding generated using the Dynamic Memory model of Doc2Vec.  The similarity score of MFeEmb accurately predicts higher classification accuracy. `*' denotes the logistic regression results, and `+' denotes the SVM results. 
}
\label{similarity}
\end{table}
\subsection{MFeEmb Performance Summary} 
Figure \ref{fig:TableForDRGita_2} provides the overall comparison of MFeEmb vs.\ the benchmark embeddings.  In the case where minimal domain adaptation was required (testing classifiers on Teams2 that were trained on Teams1), the simple version of MFeEmb using a SVM classifier is the top performer and outperforms the embeddings used in other prior work on conflict prediction~\cite{enayet2021analyzing}. Our most consistent model, MFeEmb with FsText, had a significantly higher F1 score on the high conflict class compared to baseline models (see Table \ref{classF1Scores}). Note that detecting the high conflict examples is more valuable for practical implementations.

For the more complex domain adaptation scenarios (GitHub and ASIST), the best performance was achieved using MFeEmb as a replacement for the Word2Vec embedding during the meta-training phase of FsText on GitHub, and the concatenation ensemble showed significantly better performance on the ASIST dataset. The vanilla MFeEmb generally performed comparably to the concatenation ensemble using LSTMs on out of domain datasets. The latter showed a high standard deviation compared to the former. 

To analyze the importance of incorporating word embedding in MFeEmb, we compare the performance of all the experimental setups with and without word embedding (WE). For SVM \& Logistic regression (Basic) and FsText, we train the model on the Teams Game1 dataset, and for concatenation ensemble, we train on the synonym replaced dataset. One of our main objectives in incorporating the word embedding in MFeEmb is to maintain the performance on the in-domain dataset, and results show that MFeEmb performed better with word embedding on the in-domain dataset. For most transfer case setups, MFeEmb with word embedding either gave better or comparable mean F1 scores (Figure \ref{fig:ImportanceOfWordEmb}).   The following sections present a more in-depth evaluation of each experimental setup.  
\begin{table}[!h]
\centering
\captionsetup{width=\linewidth}
\begin{tabular}{|p{3.6cm}|p{1.4cm}|p{1.4cm}|}
\hline
\multicolumn{ 3}{|c|}{\textbf{\small High Conflict Class Prediction Summary}} \\ \hline
\multicolumn{1}{|c|}{\textbf{\small Method}}&\multicolumn{1}{|c|}{\textbf{\small GitHub}}&\multicolumn{1}{|c|}{\textbf{\small ASIST}}\\\hline
\textrm{\small BERT\_SynReplace} &\small 0.431&\small 0.347
\\ \hline
\textrm{\small DA\_only\_Team1}&\small 0.320*&\small 0.311*\\ \hline
\textrm{\small Senti\_only\_Team1} &\small 0.207*&\small 0.300*\\ \hline
\textrm{\small MFeEmb\_FsText\_Team1} &\textbf{\small 0.564}&\textbf{\small 0.478}
\\ \hline
\end{tabular}
\caption{Summary of high conflict class F1 scores. 
 `*' denotes the logistic regression results, and `+' denotes the SVM results. }
\label{classF1Scores}
\end{table}
\subsection{SVM and Logistic Regression}\label{result1}
Table~\ref{MFeEmb_combined} gives the results for the SVM and logistic regression classifiers. This paper presents a thorough evaluation of the performance of different embedding choices (DM, DBOW).
We also evaluate the performance of different data augmentation methods (\textbf{SynReplace}, \textbf{ASISTReplace}, and \textbf{GitReplace}). 

Our proposed MFeEmb trained using Doc2Vec and classified using either SVM or logistic regression performed better than the word-embedding baseline. Leveraging synthetic datasets yielded significant performance improvements.  In our most challenging resource-scarce scenario, where we trained the model only on the Teams Game1 dataset, incorporating word embedding showed better performance on the Teams Game2 and GitHub datasets, while the model performed better on the ASIST dataset without word embedding (see  Figure \ref{fig:ImportanceOfWordEmb}). 
\begin{table*}[!h]
\centering
\captionsetup{width=12.5cm}
\begin{tabular}{|p{4cm}|p{2.5cm}|p{2.5cm}|p{2.5cm}|}
\hline
\multicolumn{ 4}{|c|}{\textbf{\small SVM \& Logistic Regression Results}} \\ \hline
\textbf{\small Method} & \textbf{\small Teams Game2}\newline \small F1\_score (std) & \textbf{\small GitHub}\newline \small F1\_score (std) & \textbf{\small ASIST}\newline \small F1\_score (std) \\ \hline
\textrm{\small Baseline Doc2Vec\_dbow} & \small 0.465 (0.070)*
 &\small  0.489 (0.080)* 
& \small 0.425 (0.091)* 
\\ \hline
\textrm{\small MFeEmb\_Team1\_dbow} &\small  0.533 (0.068)*
 & \small 0.437 (0.025)* 
& \small 0.347 (0.002)* 
\\ \hline
\textrm{\small MFeEmb\_Team1\_dm} &\small 0.625 (0.0295)+
 & \small 0.495 (0.012)+
& \small 0.473 (0.023)+
\\ \hline
\textrm{\small MFeEmb\_SynReplace} &\small 0.558(0.035)+ 
 &\small 0.296(0.025)* 
 &\small 0.318 (0.00)* 
\\ \hline
\textrm{\small MFeEmb\_GitReplace}& \small \textbf{0.676 (0.033)+}
& \small 0.409 (0.039)* 
& \small 0.411 (0.041)* 
 \\ \hline
\textrm{\small MFeEmb\_ASISTReplace} & \small 0.675 (0.041)+ 
 & \small \textbf{0.537 (0.060)*} 
 & \small \textbf{0.480 (0.042)*} 
\\ \hline
\multicolumn{ 4}{|c|}{\textbf{\small Concatenation Ensemble Results}} \\ \hline
\textrm{\small Baseline\_SynReplace} & \small 0.435 (0.048)
 & \small 0.414 (0.104)
 & \small 0.397 (0.081)
\\ \hline
\textrm{\small MFeEmb\_SynReplace} & \small 0.453 (0.044)
 & \small 0.429 (0.122)
 & \small 0.459 (0.044)
\\ \hline
\textrm{\small MFeEmb\_GitReplace}& \textbf{\small 0.464 (0.044)}
& \small 0.468 (0.098)
& \textbf{\small 0.491 (0.054)}
 \\ \hline
\textrm{\small MFeEmb\_ASISTReplace} & \small 0.408 (0.075)
 & \textbf{\small 0.516 (0.100)}
 & \small 0.455 (0.059)
\\ \hline
\multicolumn{ 4}{|c|}{\textbf{\small Few-Shot Learning Results}} \\ \hline
\textrm{\small FsText Baseline} & \textbf{\small 0.689 (0.0)}
 & \small 0.330 (0.0)
 & \small 0.338 (0.0)
 \\ \hline
\textrm{\small MFeEmb\_Team1\_doc2Vec} & \small 0.60 (0.028)
 & \small 0.583 (0.045)
 & \small 0.451 (0.025)
\\ \hline
\textrm{\small MFeEmb\_Team1\_word2Vec} & \small 0.597 (0.041)
 & \small 0.507 (0.063)
 & \small 0.437 (0.027)
\\ \hline
\textrm{\small MFeEmb\_SynReplace} & \small 0.544 (0.021)
 & \small 0.568 (0.031)
 & \small 0.435 (0.037)
\\ \hline
\textrm{\small MFeEmb\_GitReplace}& \small 0.684 (0.033)
& \small 0.567 (0.041)
& \small 0.388 (0.266)
 \\ \hline
\textrm{\small MFeEmb\_ASISTReplace} & \small 0.664 (0.042)
 & \textbf{\small 0.608 (0.034)}
 & \textbf{\small 0.462 (0.053)}
\\ \hline
\end{tabular}
\caption{\small Detailed performance evaluation of MFeEmb.  `*' denotes the logistic regression results, and `+' denotes the SVM results.} 
\label{MFeEmb_combined}
\end{table*}
\subsection{Concatenation Ensemble Model}\label{result2}
Table \ref{MFeEmb_combined} gives the results for the LSTM-based concatenation ensemble model. The model showed a better mean F1-score than the text-based LSTM model. We also trained the LSTM using synthetic datasets generated using GitHub and ASIST vocabularies, which showed better performance, specifically with the GitHub vocabulary dataset. The model performed significantly better on the ASIST dataset compared to the other experimental setups. 
\subsection{Few-Shot Model (FsText)}\label{result3}
The FsText baseline showed the best performance on Game2, but the performance degraded considerably on the transfer task (GitHub and ASIST). FsText with the proposed MFeEmb exhibited significantly better performance on the GitHub and ASIST datasets, specifically with ASIST vocabulary's synthetic dataset. FsText with the proposed MFeEmb embedding also gave a comparable performance on the Teams Game2 dataset.  This demonstrates that MFeEmb is an excellent representation for meta-pretraining a few-shot model on collaborative multiparty dialogues even when learned from a small dataset (see Table~\ref{MFeEmb_combined}). 

Using a synthetic dataset showed a performance improvement in all three experimental setups. Generation of the synthetic dataset using the vocabulary of other collaborative tasks showed comparatively better performance on the transfer task. Even in the in-domain experiments, the Game1 Synthetic dataset, generated using collaborative task vocabulary, showed the best and comparable performance on Game2 in all the experimental setups.
\section{Conclusion and Future Work} This paper introduces a multi-feature embedding (MFeEmb) to improve the generalizability of multiparty dialogue models under resource scarcity scenarios.   We propose the use of a combination of textual (words), structural (DAs), and semantic (sentiment, DAs) embeddings to reduce the performance loss due to natural distribution shift.  Experiments show that the multi-feature embedding performs significantly better than sentence (BERT), dialogue act-only, sentiment-only, and word embeddings.  Our results demonstrate that MFeEmb is a superior representation for meta-pretraining a few-shot model that works well across different collaborative problem-solving domains. 
 
 Our proposed data augmentation strategy successfully resolved the domain shift problem caused by task-specific vocabulary without perturbing the dialogue act and sentiment features.  
Experiments with synthetic datasets show that synonym replacement with vocabulary drawn from a collaborative task outperforms generic synonym replacement with WordNet. It improves both the transfer accuracy and the test accuracy on the in-domain test set. Note that we did not fine-tune the models on the target datasets, i.e., GitHub and ASIST, and strictly report the model learned on the Teams dataset.
Only the vocabulary of these datasets was used to boost the performance; explicit fine-tuning of the machine learning models could further improve the results. 

\section{Limitations} This paper only reports results on the generalizability of MFeEmb on conflict prediction tasks; MFeEmb may not perform as well on other communication analysis tasks. However, we believe that modifying the features used in the embedding can address this problem. In future work, we are interested in applying our embedding to new team communication analysis tasks such as identifying emergent leadership.

\section{Acknowledgments}
This material is based upon work supported by the Defense Advanced Research Projects Agency (DARPA) under Contract No. W911NF-20-1-0008 and the National Institute of Mental Health of the National Institutes of Health under Award Number P50MH126231.  Any opinions, findings and conclusions or recommendations expressed in this material are those of the authors and do not necessarily reflect the views of DARPA, the National Institutes of Health, and  the University of Central Florida.

\bibliography{EACL_Prep_CameraReady}
\bibliographystyle{acl_natbib}
\newpage
\appendix
\begin{figure*}
    \centering
    \captionsetup{width=.9\linewidth}
    \includegraphics[width=0.90\textwidth]{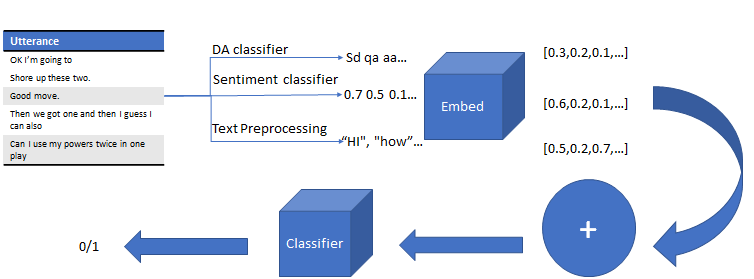}
    \caption{Utterances are classified using the dialogue act classifier to produce a sequence of DAs and the sentiment classifier to produce a time series of sentiment polarities. Along with the text data, these sequences are used to create MFeEmb using the Dynamic Memory model of Doc2Vec. The few shot learning and data augmentation options are not shown in the figure.}
    \label{fig:MFeEmb}
\end{figure*}

\section{Dialogue Act Classification} 
\label{App_DAC}We use a dialogue act classifier (USE-DAC) to map dialogues to the sequence of DAs, where each DA in a sequence corresponds to the utterance of the dialogue. Utterances are tagged according to the SwDA-DAMSL tagset\footnote{\url{https://web.stanford.edu/~jurafsky/ws97/manual.august1.html}} which contains 42 tags, and one sequence is generated per dialogue.  Our dialogue act classifier uses the Universal Sentence Encoder (USE) module available at TensorFlow Hub\footnote{ \url{https://tfhub.dev/google/universal-sentence-encoder-large/2}}.  After extensive experiments, we identified that USE with three dense layers performs best on transfer tasks. We selected the USE Transformer-based Architecture model with three dense
layers and a softmax activation function. We fine-tune USE on the SwDA dataset and use the classifier to tag the utterances of the test and training datasets.  We selected the USE transformer-based model because it is itself trained on dialogue and discussion forum datasets. The test accuracy of the classification model is 72\%.

\section{Doc2Vec} \label{App_doc2vec}Doc2Vec~\cite{docvec} is an unsupervised method to learn paragraph vectors from text of arbitrary size. We represent each dialogue as 1) sequence of utterances, 2) sequence of DAs, and 3) sentiment polarities. We pass these sequences through Doc2Vec to generate representations. 
We use the Doc2Vec implementation from the python Gensim library with an epoch size of 5, negative sampling 5, window size 5, and alpha 0.065.
\section{SVM \& Logistic Regression} We use the classifier implementations from the scikit-learn library. The SVM was trained using the RBF kernel function and the default parameters. The parameters for logistic regression were:
Cs=10, class\_weight=None, cv=10, dual=False,
           fit\_intercept=True, intercept\_scaling=1.0, max\_iter=1000,
           multi\_class='ovr', n\_jobs=None, penalty='l2', random\_state=5434,
           refit=True, scoring='accuracy', solver='lbfgs', tol=0.001,
           verbose=False.
Table \ref{SvmLg_both} shows the full results of both the SVM and Logistic Regression classifiers.           

\section{BERT Baseline}
\label{sec:bert}
We use the bert\_en\_uncased\_L\-12\_H\-768\_A\-12 model available at TensorFlow Hub\footnote{\url{https://tfhub.dev/tensorflow/bert_en_uncased_L-12_H-768_A-12/4}} to develop our baseline classifier. The model contains one dense layer, one dropout layer, sigmoid activation function, Adam optimizer. Due to the small size of the Game1 dataset we train the model on the synonym replaced Game1 dataset. The total number of parameters in the model is: 10,948,301.
\section{FsText} The original FsText works by using the pre-trained word2Vec embedding model word2vec-google-news-300 available through the gensim.downloader module. For MFeEmb we generate the embedding using Doc2Vec with the same parameters mentioned in Appendix~\ref{App_doc2vec}. The second phase uses a cosine similarity-based classification model that does not involve machine learning.  
\section{Concatenation Ensemble}
Our Bidirectional LSTM models for each feature has an embedding layer, an LSTM layer, one dropout layer, and one deep layer. The LSTM uses a sigmoid activation function and is trained using the Adam optimizer with a learning rate=0.01.  The output shape of each individual model is (None, 100). The total  number of parameters of the concatenation ensemble is: 2,365,081.

\section{GitHub Dataset Conflict Scoring}
\label{sec:github_scoring}
For GitHub, process conflict was scored according to issue resolution using the following heuristics to determine if conflicts occurred: 
\begin{compactenum}
    \item Unsuccessful resolution of the issue. 
    \item Unanswered questions in the discussion. 
    \item Lack of understanding about the issue from one or more members.
    \item Lack of understanding or disagreement between the team members.
    \item Disagreement between the members about the proposed solution.
\end{compactenum}

\section{Dialogue Act Frequency Distribution Analysis} 
\label{sec:freq_distrib}
\begin{figure*}[!h]
    \centering
    \captionsetup{width=.9\linewidth}
    \includegraphics[width=0.90\textwidth]{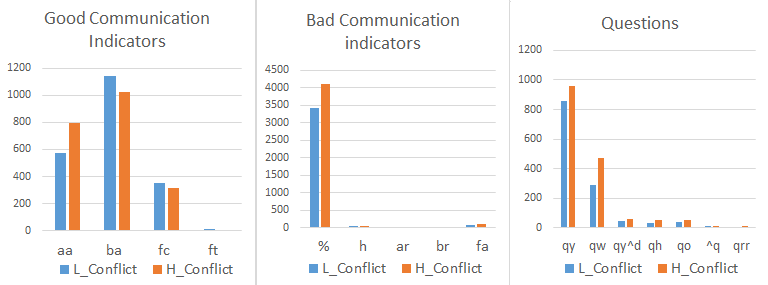}
    \caption{Dialogue act frequency distribution in high and low conflict classes for the Teams dataset.  Dialogue acts were divided into good and bad communication indicators.}
    \label{fig:DA_feq_Distribution_1}
\end{figure*}

\begin{figure*}[!h]
    \centering
    \captionsetup{width=.9\linewidth}
    \includegraphics[width=0.90\textwidth]{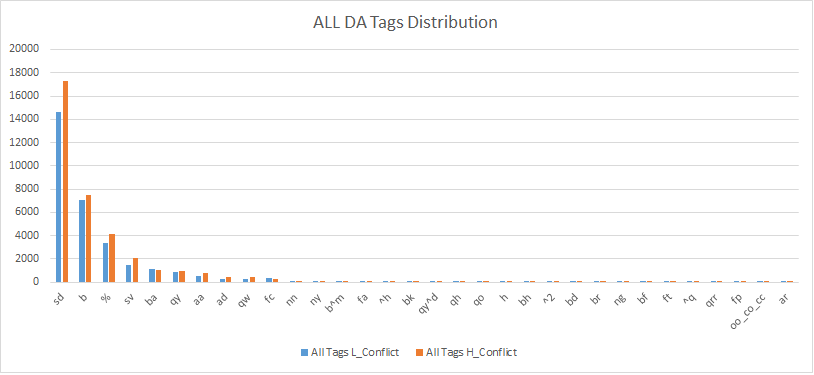}
    \caption{Complete dialogue act frequency distribution for high and low conflict classes in the Teams dataset.}
    \label{fig:All_DA_distribution_1}
\end{figure*}

\begin{figure*}[!h]
\captionsetup{width=\linewidth}
    \centering
    \includegraphics[width=0.9\textwidth]{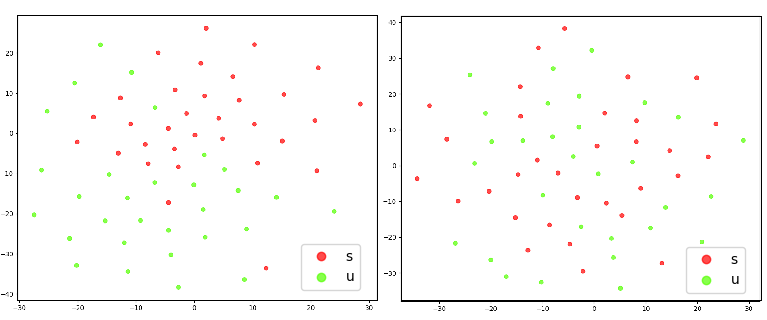}
    \caption{Comparison of the MFeEmb and word embedding distribution on the 2D plane. Multi-feature embedding showed better clustering, with most instances of one of the classes occupying the lower left and the other occupying the upper right. On the other hand, word embeddings are very intermixed. s: low conflict (successful dialogue), u: high conflict (unsuccessful dialogue).}
    \label{fig:embdDist}
\end{figure*}

Figure~\ref{fig:DA_feq_Distribution_1} shows the frequency distribution for dialogue acts in the low conflict and high conflict classes of the Teams dataset.  We divided the dialogue acts into good communication indicators, bad communication indicators and questions.  Figure~\ref{fig:All_DA_distribution_1} shows the complete dialogue act frequency distribution. Table~\ref{tab:grams} shows the n-gram frequency distribution of dialogue acts across all datasets.  Figure~\ref{fig:embdDist} visualizes the separation between low conflict and high conflict classes using both the MFeEmb and word embedding.  To see if profanities were a reliable indication of conflict, we also examined profanity vocabulary differences.
The most frequent words in the high conflict dialogues that are in profanity list are:
['hell', 'kill','suck','sucking','shit','strip','stroke', 'rectum','xxx','dick','screwed','retard', 'ovary','piss','lube', 'junkie'].

The most frequent words in the low conflict dialogues that are in the profanity list are:
['booty', 'pot','carpet', 'rum', 'breasts', 'pedophile', 'urine', 'thug', 'screw', 'jerk', 'weed', 'screwing', 'shower', 'stupid'].


\begin{table*}
\centering
\begin{tabular}{p{1cm}p{2cm}p{3cm}p{2cm}p{2.3cm}p{2.6cm}}
\hline
\textbf{Dataset} & \textbf{Unigrams}& \textbf{Bigrams}& \textbf{Trigrams}& \textbf{4grams}& \textbf{5grams}\\
\hline
Teams&(sd),(b),(\%) & (sd,sd),(sd,b),(b,sd) & (sd,sd,sd),\newline(sd,sd,b),\newline(sd,b,sd) & (sd,sd,sd,  sd),\newline (sd,sd,sd,sd), \newline(sd,sd,sd,b) & (sd,sd,sd,sd, sd), \newline(sd,sd,sd,sd,b),\newline(sd, sd,sd,b,sd)\\
\hline
GitHub&(sd),(sv),(ad) &	(sd,sd),(sd,sv),(sv,sd)&(sd,sd,sd),\newline(sv,sd,sd),\newline(sd,sd,ad)&	(sd,sd,sd,sd),\newline(sd,sd,sd,ad), \newline(sv,sd,sd,sd)&(sd,sd,sd,sd,sd),\newline(sd,sd,sd,sd, ad),\newline(sd,sv,sd,sd,sd)\\
\hline
ASIST&	(sd),(qy),(sv)&	(sd,sd),(sd,qy),(qy,sd)& (sd,sd,sd),\newline(sd,qy,sd),\newline(qy,sd, sd)&(sd,sd,sd,sd), \newline(sd,qy,sd,sd), \newline(sd,sd,qy,sd)&(sd,sd,sd,sd,sd), \newline(sd,sd,sd,sd, qy), \newline(sd,qy,sd,sd,sd),
\\
\hline
\end{tabular}
\caption{N-gram frequency distribution: top three most frequent unigrams, bigrams, trigrams, 4grams, 5grams of all the datasets.  Sequences of sd (statement-nonopinion) are common across all datasets.}
\label{tab:grams}
\end{table*}

\section{Synthetic Datasets}
Synthetic datasets were generated using: \url{https://github.com/jasonwei20/eda_nlp}.


\section{Results on Adversarially Generated Dataset}
\label{app:adversary}
This section presents results on the adversarially generated dataset (Synthetic Game 2) created using TextAttack\footnote{\url{https://github.com/QData/TextAttack}}. Word Swap by BERT-Masked LM transformation was employed to generate synthetic examples from the Teams Game2 dataset. One synthetic example is generated per dialogue of the Game2 dataset.  The length of the synthetic Game2 dataset vocabulary is 6084, and the length of the original Game1 dataset vocabulary is 3441. The number of words in the synthetic dataset that are not in the original Game1 is 3904.

Figure \ref{fig:ven_diagram} shows a high overlap between original Game1 and original Game2 compared to synthetic Game2 and original Game1, but this does not affect the performance of MFeEmb (Basic), and MFeEmb. (Basic) gave a better performance on the synthetic dataset. On the other hand, the performance of the BERT baseline decreased on the synthetic Game2 test set, with a high standard deviation in mean F1 scores.

\begin{table*}[!h]
\centering
\begin{tabular}{|p{3cm}|p{2.5cm}|p{2.5cm}|p{2.5cm}|p{2.5cm}|}
\hline
\multicolumn{ 5}{|c|}{\textbf{Game2 Synthetic Dataset Results}} \\ \hline
\textbf{Train model} & \textbf{Teams Game1}\newline F1\_score (std) & \textbf{SynReplace}\newline F1\_score (std)&\textbf{GitReplace}\newline F1\_score (std) & \textbf{ASISTReplace}\newline F1\_score (std) \\ \hline
MFeEmb (Basic)
& 0.654 (0.033)+
 & 0.443 (0.046)*
& \textbf{0.617 (0.035)+}
&\textbf{0.624 (0.055)+}
\\ \hline
BERT
& -
 & \textbf{0.490 (0.061)}
& 0.422 (0.037)
& 0.495 (0.044)
\\ \hline
\end{tabular}
\caption{\small MFeEmb results on the Game2 synthetic dataset generated using TextAttack}
\label{Sig_Biclas_SynthticGame2}
\end{table*}
\section{SVM \& Logistic Regression Results}\label{sec:svmLogresults}
Table \ref{SvmLg_both}, \ref{similarity1}, \ref{classF1Scores1} provides the detailed results of both the SVM and logistic regression classifiers under different experimental settings.  
\begin{table*}
\centering
\begin{tabular}{|p{4.5cm}|p{3cm}|p{3cm}|p{3cm}|}
\hline
\multicolumn{ 4}{|c|}{\textbf{Both SVM \& Logistic Regression Results}} \\ \hline
\textbf{Method} & \textbf{Teams Game2}\newline F1\_score (std) & \textbf{GitHub}\newline F1\_score (std) & \textbf{ASIST}\newline F1\_score (std) \\ \hline
\textrm{\small Baseline Doc2Vec\_dbow} & 0.465 (0.070)*\newline 0.369 (0.0)+
 & 0.489 (0.080)* \newline 0.425 (0.0)+
& 0.425 (0.091)* \newline 0.348 (0.0)+
\\ \hline
\textrm{\small MFeEmb\_Team1\_dbow} & 0.533 (0.068)*\newline 0.369 (0.0)+
 & 0.437 (0.025)* \newline 0.425 (0.0)+
& 0.347 (0.002)* \newline 0.348 (0.0)+
\\ \hline
\textrm{\small MFeEmb\_Team1\_dm} & 0.625 (0.0295)+ \newline 0.569 (0.045)*
 & 0.495 (0.012)+ \newline 0.428 (0.0)*
& 0.473 (0.023)+ \newline 0.393 (0.032)*
\\ \hline
\textrm{\small MFeEmb\_SynReplace} & 0.558(0.035)+ \newline 0.542 (0.045)*
 & 0.296(0.025)* \newline 0.248 (0.0)+
 & 0.318 (0.00)* \newline 0.318 (0.00)+
\\ \hline
\textrm{\small MFeEmb\_GitReplace}& \textbf{0.676 (0.033)+} \newline 0.593 (0.056)*
& 0.409 (0.039)* \newline 0.248 (0.0)+
& 0.411 (0.041)* \newline 0.318 (0.00)+
 \\ \hline
\textrm{\small MFeEmb\_ASISTReplace} & 0.675 (0.041)+ \newline 0.643 (0.044)*
 & \textbf{0.537 (0.060)*} \newline 0.248 (0.0)+
 & \textbf{0.480 (0.042)*} \newline 0.318 (0.00)+
\\ \hline
\end{tabular}
\caption{\small Results for both the SVM and logistic regression classifiers side by side.}
\label{SvmLg_both}
\end{table*}

\begin{table}[!h]
\centering
\captionsetup{width=\linewidth}
\begin{tabular}{|p{1.5cm}|p{1.5cm}|p{1.5cm}|p{1.5cm}|}
\hline
\multicolumn{2}{|c|}{\textbf{\small Word\_{Emb}}}&\multicolumn{2}{|c|}{\textbf{\small MFeEmb}}\\
\hline
\multicolumn{4}{|c|}{\textbf{\small Teams Game2}}\\
\hline
\small similarity&\small F1\_score&\small similarity&\small F1\_score\\
\hline
\small -0.067&\small 0.470*\newline 0.369+&\small -0.016&\textbf{\small 0.628+}\newline \small 0.561*\\
\hline
\multicolumn{4}{|c|}{\textbf{\small GitHub}}\\
\hline
\small similarity&\small F1\_score&\small similarity&\small F1\_score\\
\hline
\small -0.067&\small 0.463*\newline 0.425+&\small -0.017&\textbf{\small 0.501+}\small \newline 0.439*\\
\hline
\multicolumn{4}{|c|}{\textbf{\small ASIST}}\\
\hline
\small similarity&\small F1\_score&\small similarity&\small F1\_score\\
\hline
\small -0.067&\small 0.446+\newline 0.348*&\small -0.016&\textbf{\small 0.458+}\small \newline 0.394*\\
\hline
\end{tabular}
\caption{Similarity-based generalizability analysis. '*' denotes the logistic regression results, and '+' denotes the SVM results.}
\label{similarity1}
\end{table}

\begin{table}[!h]
\centering
\captionsetup{width=\linewidth}
\begin{tabular}{|p{3.6cm}|p{1.4cm}|p{1.4cm}|}
\hline
\multicolumn{ 3}{|c|}{\textbf{\small High Conflict Class Prediction Summary}} \\ \hline
\multicolumn{1}{|c|}{\textbf{\small Method}}&\multicolumn{1}{|c|}{\textbf{\small GitHub}}&\multicolumn{1}{|c|}{\textbf{\small ASIST}}\\\hline
\textrm{\small BERT\_SynReplace} &\small 0.431&\small 0.347
\\ \hline
\textrm{\small DA\_only\_Team1}&\small 0.320*\newline 0.250+&\small 0.311*\newline0.216+\\ \hline
\textrm{\small Senti\_only\_Team1} &\small 0.207*\newline0.090+&\small 0.300*\newline 0.036+\\ \hline
\textrm{\small MFeEmb\_FsText\_Team1} &\textbf{\small 0.564}&\textbf{\small 0.478}
\\ \hline
\end{tabular}
\caption{Summary of high conflict class F1\_scores.'*' denotes the logistic regression results, and '+' denotes the SVM results. }
\label{classF1Scores1}
\end{table}

\end{document}